\documentclass{article}

\usepackage{PRIMEarxiv}
\usepackage[utf8]{inputenc}
\usepackage[T1]{fontenc}
\usepackage{hyperref}
\usepackage{url}
\usepackage{booktabs}
\usepackage{amsfonts}
\usepackage{nicefrac}
\usepackage{microtype}
\usepackage{amsmath}
\usepackage[inline]{enumitem}
\usepackage{algorithm}
\usepackage{algpseudocode}
\usepackage{rotating}
\usepackage{multirow}
\usepackage{makecell}
\usepackage{threeparttable}
\usepackage{graphicx}
\usepackage{subcaption}
\graphicspath{{media/}}
\usepackage[title]{appendix}

\title{HiCal: Offline Calibration for Hierarchical Scale Encoding in Accurate HiF4 Quantization}

\author{
Zhiyuan Zhang, Yanzhao Li, Hui Dong, Sifan Cheng, Mingen Zheng, Hui Wang \\
Huawei Technology \\
\texttt{\{zhangzhiyuan45, liyanzhao\}@huawei.com}
}

\begin{document}

\maketitle


\begin{abstract}
Block floating-point (BFP) quantization is widely adopted for efficient low-bit large language model (LLM) inference, where the choice of block scales plays a critical role in quantization accuracy. HiFloat4 (HiF4) employs a hierarchical scale encoding scheme consisting of a global E6M2 scale and two levels of binary scale extensions. However, its conventional rounding rule and fixed scale-extension thresholds may not fully capture the statistical characteristics of the quantized tensors in practice, leading to suboptimal scale representations.
In this work, we propose HiCal, an offline calibration approach that
improves HiF4 quantization accuracy by refining the encoding decisions
across all three scale levels. Specifically, we introduce a calibrated rounding bias for the Level-1 E6M2 scale and optimize the scale-extension thresholds at Levels 2 and 3 based on the empirical distributions of local group magnitudes. The resulting calibration parameters are determined offline and remain fixed during inference, while the hierarchical encoding continues to adapt to individual blocks through their local values. This design eliminates the need for per-block multi-candidate scale search while preserving the original deterministic HiF4 encoding procedure.
Experiments across five synthetic and real activation distributions show that the proposed method reduces quantization MSE by 9.91\%–23.50\% relative to the original HiF4 scheme, including a 13.00\% reduction on real Qwen3-8B activations. End-to-end evaluations across six benchmarks further show average-score improvements of 0.91 and 1.25 percentage points on Qwen3-8B and Qwen3-30B-A3B, respectively, without per-block multi-candidate search during inference.
\end{abstract}

\section{Introduction}
\label{sec:intro}

The growing computational and memory demands of large language models
(LLMs) have made low-bit quantization an important technique for
efficient model inference. Recent advances in low-precision numerical
formats, including FP8 and FP4 representations, have enabled substantial
reductions in model storage and data-transfer costs
\cite{micikevicius2022fp8,liu2023llmfp4}. In particular, block floating-point
(BFP) and microscaling formats, such as MXFP4 and NVFP4, use shared
scaling metadata to represent groups of low-precision elements
\cite{ocp_mx,nvfp4}. These formats provide a practical way to extend
the numerical range of low-bit representations while maintaining
hardware-friendly computation. However, at the 4-bit level, the
quantization accuracy becomes highly sensitive to the selection and
encoding of scaling factors, since even small changes in the scale
can substantially affect the reconstructed values within a block.

Conventional BFP quantization generally initializes the block scale
according to the maximum absolute value of the block. Although this
max-based strategy prevents overflow and fully utilizes the available
quantization range, it does not necessarily minimize the overall
quantization--dequantization (QDQ) error. This limitation has motivated
recent studies on more effective block-scale selection strategies.
Four Over Six introduces an adaptive scaling method that evaluates
two candidate scale factors for each block and selects the one that
produces lower quantization error
\cite{cook2025fouroversix}. ScaleSearch further explores fine-grained
search over neighboring representable scale values to identify scales
that minimize block-level reconstruction error
\cite{gupta2026scalesearch}. More recently, SOAR proposes analytical
optimization of global and block-wise scales, together with a
decoupled scale-search strategy, while H-Scale uses a diagonal
second-order approximation derived from calibration activations to
refine per-group scales
\cite{bao2026soar,yu2026hscale}. These methods demonstrate that
optimizing the scale-selection process beyond conventional max-based
initialization can improve low-bit quantization accuracy, while
introducing different optimization objectives and calibration
procedures. In addition to scale selection, activation outliers remain a major
obstacle for low-bit microscaling formats. MX+ addresses this by
retaining outlier information outside the standard BFP representation
\cite{lee2025mxplus}, while AMXFP4 uses an asymmetric FP4 format to
absorb activation outliers without calibration \cite{kim2025amxfp4}.
Channel-wise outlier separation has also been explored for W4A4
quantization \cite{zhang2026osc}.

Other studies have investigated the interaction between microscaling
granularity, scale representation, and quantization accuracy, and
have examined how activation outliers interact with these choices
\cite{lee2025mxplus,kim2025amxfp4,zhang2026osc}.
For example, recent analyses show that reducing the block size does
not always improve model accuracy, as the resulting behavior depends
on the statistical properties of the tensor distribution and the
dynamic range of the scaling format
\cite{fasoli2026finer,schaefer2026finer}. In addition, format-specific
post-training quantization methods, such as Micro-Rotated-GPTQ,
combine block-wise Hadamard transformations with FP4-oriented
optimization to address the limitations of existing microscaling
formats
\cite{egiazarian2026bridging}. These studies highlight that low-bit
quantization accuracy depends not only on the element-level codebook,
but also on the interaction among scale representation, scaling
granularity, and the encoding procedure.

HiFloat4 (HiF4) is a 4-bit block floating-point format that employs
a three-level hierarchical scaling structure to capture magnitude
variations at different granularities
\cite{luo2026hifloat4}. As illustrated in Fig.~\ref{fig:HiF4_s1p2},
each HiF4 block contains 64 four-bit S1P2 elements and 32 bits of
shared scaling metadata, corresponding to an average storage cost
of 4.5 bits per element. The metadata consists of one global E6M2
base scale at Level 1, eight 1-bit scale extensions
($\mathrm{E}_8$) at Level 2, and sixteen 1-bit scale extensions
($\mathrm{E}_4$) at Level 3. The two levels of scale extensions
provide progressively finer adjustments for groups of eight and
four elements, respectively. Accordingly, the effective scale
of the $i$-th element can be expressed as
\begin{equation}
    S_i =
    S_{\mathrm{base}}
    \cdot
    2^{
        E_8[\lceil i/8\rceil]
        +
        E_4[\lceil i/4\rceil]
    },
    \quad i=1,\ldots,64,
\end{equation}
where $S_{\mathrm{base}}$ denotes the E6M2 base scale, and
$e^{(8)}$ and $e^{(4)}$ represent the corresponding binary
scale extensions. In the original HiF4 encoding procedure,
the base scale is initialized through max-based normalization
and E6M2 encoding, while the Level-2 and Level-3 scale extensions
are determined by fixed magnitude thresholds. This hierarchical
design provides flexible local scale adaptation with compact
metadata, but its final quantization accuracy also depends on
how the encoding decisions at the three levels are configured.

A key characteristic of HiF4 is that its three scale levels are
coupled. The Level-1 E6M2 base scale determines the normalization
of local group magnitudes, which subsequently affects the
Level-2 and Level-3 scale-extension decisions. Therefore, changing
the Level-1 rounding rule may alter not only the global scale
but also the assignment of the two subsequent levels of scale
extensions. Meanwhile, the two magnitude thresholds that decide the Level-2 and Level-3 scale-extension bits may not be optimal for the tensors being quantized. These observations motivate a systematic investigation of the scale-encoding decisions across the complete
HiF4 hierarchy, rather than optimizing the element-level codebook
or an individual scale factor in isolation.

In this work, we propose HiCal, a lightweight, offline-calibrated
approach to improve the scale encoding accuracy of HiF4.
Instead of performing multi-candidate scale evaluation independently
for every block during quantization, we calibrate a compact set of shared encoding parameters offline,
following the offline-calibration paradigm of prior post-training
quantization methods \cite{xiao2023smoothquant}, and apply them as
fixed rules. Specifically, we introduce a rounding bias for the Level-1 E6M2
base scale and recalibrate the scale-extension thresholds at
Levels 2 and 3 according to the empirical distributions of local
group magnitudes. The calibration is performed sequentially:
the Level-1 rounding bias is first selected, and the Level-2 and
Level-3 thresholds are subsequently calibrated with the selected
bias fixed. The resulting configuration uses
$\alpha_{\mathrm{rnd}}=-0.35$, $TH_8=4.25$, and $TH_4=2.125$.
During quantization, the selected parameters are directly applied
to each block, while the actual hierarchical scale metadata
continues to be determined from its local values. Thus, the
proposed approach retains the original HiF4 hierarchical
representation and S1P2 element codebook, without requiring
per-block multi-candidate scale evaluation.

We evaluate the proposed method using five synthetic and real-world activation distributions for scale-encoding analysis, and conduct end-to-end experiments on Qwen3-8B and Qwen3-30B-A3B in which both weight and activation tensors are quantized with the same HiF4 format. The results show
that the proposed calibration reduces QDQ mean squared error
(MSE) by 9.91\%--23.50\% relative to the original HiF4 scheme,
including a 13.00\% reduction on real Qwen3-8B activations.
Furthermore, evaluations on six downstream benchmarks show
that the six-task mean accuracy improves by 0.91 and
1.25 percentage points for Qwen3-8B and Qwen3-30B-A3B,
respectively.

The main contributions of this work are summarized as follows:
\begin{itemize}
    \item We characterize the hierarchical scale-encoding
    mechanism of HiF4 and analyze how the Level-1 base-scale
    rounding and the Level-2/Level-3 scale-extension decisions
    affect the final QDQ error.

    \item We propose a lightweight, offline-calibrated
    approach that sequentially determines the Level-1 E6M2
    rounding bias and the Level-2/Level-3 scale-extension
    thresholds. The calibrated parameters are fixed during
    quantization, avoiding per-block multi-candidate scale
    evaluation while preserving the original HiF4 representation.

    \item We evaluate the proposed method on five synthetic and real-world activation distributions  and on Qwen3-8B and Qwen3-30B-A3B. The results demonstrate consistent QDQ
    error reductions and improved average downstream
    benchmark accuracy.
\end{itemize}

\section{HiF4 Scale Encoding and Motivation}
\label{sec:background}
\subsection{HiF4 Architecture and Scale-Encoding Decisions}
\label{subsec:HiF4_quantization}

\begin{figure}[t]
    \centering
    \includegraphics[width=0.77\linewidth]{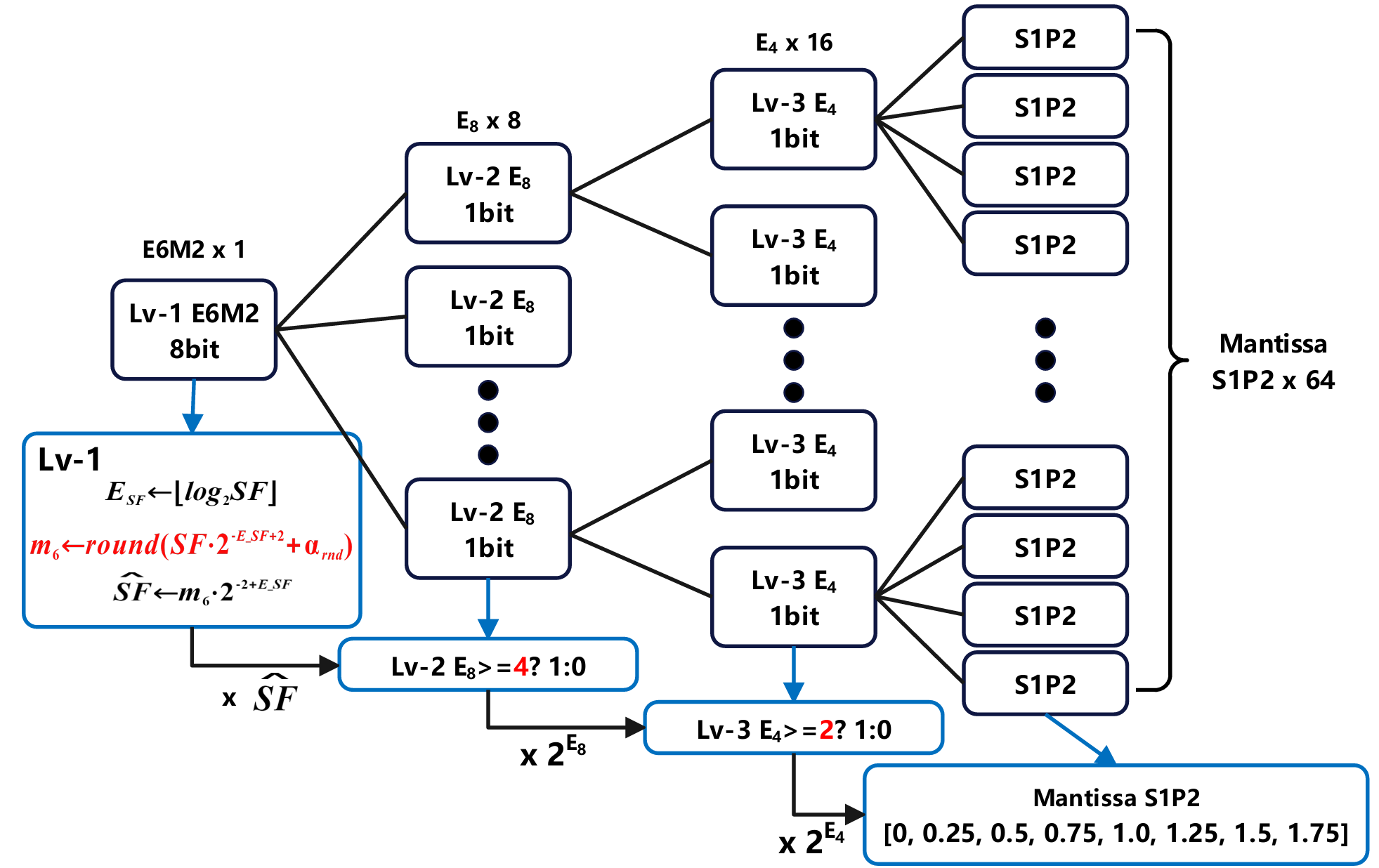}
    \caption{HiF4 three-level hierarchical scaling architecture.
    Each 64-element block shares one E6M2 global scale factor
    ($\mathrm{SF}$), eight 1-bit Level-2 scale extensions
    ($\mathrm{E}_8$), and sixteen 1-bit Level-3 scale extensions
    ($\mathrm{E}_4$), providing scale adjustments at the block,
    eight-element-group, and four-element-group granularities,
    respectively. The three components highlighted in red---the E6M2
    rounding rule and the Level-2 and Level-3 magnitude thresholds---are
    the parameters calibrated in this work; all other operations follow
    the original HiF4 procedure.}
    \label{fig:HiF4_s1p2}
\end{figure}

HiF4 adopts a three-level hierarchical scaling architecture to represent low-precision values while capturing local magnitude variations at different granularities\cite{luo2026hifloat4}. As illustrated in Fig.~\ref{fig:HiF4_s1p2}, each HiF4 block contains 64 four-bit elements and 32 bits of shared scale metadata, consisting of one E6M2 global scale factor ($\mathrm{SF}$), eight 1-bit mid-level scale extensions ($\mathrm{E}_8$), and sixteen 1-bit fine-level scale extensions ($\mathrm{E}_4$). The three levels provide global, eight-element-group, and four-element-group scaling adjustments, respectively.
Given an input block $V_{64}=\{v_1,\ldots,v_{64}\}$, HiF4 first extracts local maximum magnitudes at two granularities. The block is partitioned into sixteen groups of four elements, and the maximum absolute value of each group is calculated as
\begin{equation}
    V_4[i]
    =
    \max_{j=4i-3,\ldots,4i}|v_j|,
    \quad i=1,\ldots,16.
    \label{eq:HiF4_v4}
\end{equation}
Two adjacent four-element groups are then combined to obtain eight-element group maxima:
\begin{equation}
    V_8[k]
    =
    \max(V_4[2k-1],V_4[2k]),
    \quad k=1,\ldots,8.
    \label{eq:HiF4_v8}
\end{equation}
The global maximum magnitude of the block is
\begin{equation}
    V_{\max}
    =
    \max_{k=1,\ldots,8}V_8[k].
    \label{eq:HiF4_vmax}
\end{equation}

\paragraph{Level-1: Global E6M2 scale encoding.}
The initial global scale is determined by the maximum magnitude of the block:
\begin{equation}
    \mathrm{SF}
    =
    \operatorname{clamp}
    \left(
        \frac{V_{\max}}{7},
        2^{-48},
        49152
    \right).
    \label{eq:HiF4_sf}
\end{equation}
The denominator $7 = 4 \cdot 1.75$ reserves headroom for the maximum scale extension ($DE=2$) plus the largest S1P2 code $1.75$, so that the largest-magnitude element remains representable under the full hierarchy. To encode the scale in E6M2, its exponent and normalized mantissa are calculated as
\begin{equation}
    E_{\mathrm{SF}}
    =
    \left\lfloor\log_2\mathrm{SF}\right\rfloor,
    \qquad
    r
    =
    \mathrm{SF}\cdot 2^{2-E_{\mathrm{SF}}}.
    \label{eq:HiF4_sf_normalization}
\end{equation}
The normalized mantissa is then rounded and mapped back to the E6M2 scale:
\begin{equation}
    m_6
    =
    \operatorname{round}
    \left(r+\alpha_{\mathrm{rnd}}\right),
    \qquad
    \widehat{\mathrm{SF}}
    =
    m_6\cdot 2^{E_{\mathrm{SF}}-2},
    \label{eq:HiF4_e6m2}
\end{equation}
where $\alpha_{\mathrm{rnd}}$ is the rounding adjustment parameter.
If the rounded mantissa falls outside $[4,7]$, the code is
canonicalized by carrying into the exponent ($m_6{=}8 \to
m_{\mathrm{can}}{=}4$, $E_{\mathrm{SF}}{+}{=}1$) or borrowing
from it ($m_6{=}3 \to m_{\mathrm{can}}{=}6$,
$E_{\mathrm{SF}}{-}{=}1$), so $\widehat{\mathrm{SF}}$ preserves the
rounded value while remaining a legal E6M2 code; the procedure is
given in Algorithm~\ref{alg:l1_bias_calibration}. The original
HiF4 encoding corresponds to $\alpha_{\mathrm{rnd}}=0$, which
applies standard rounding to the normalized E6M2 mantissa.

\paragraph{Level-2: Eight-element scale extension.}
Based on the decoded global scale $\widehat{\mathrm{SF}}$, each eight-element group is assigned a binary scale extension according to its local maximum magnitude:
\begin{equation}
    \mathrm{E}_8[k]
    =
    \mathbb{I}
    \left(
        \frac{V_8[k]}{\widehat{\mathrm{SF}}}
        \geq \mathrm{TH}_8
    \right),
    \quad k=1,\ldots,8,
    \label{eq:HiF4_e8}
\end{equation}
where $\mathrm{TH}_8=4$ in the original HiF4 configuration.

\paragraph{Level-3: Four-element scale extension.}
The Level-3 scale extension further adjusts the effective scale at the four-element-group granularity. Its binary decision is conditioned on the corresponding Level-2 extension:
\begin{equation}
    \mathrm{E}_4[i]
    =
    \mathbb{I}
    \left(
        \frac{V_4[i]}
        {\widehat{\mathrm{SF}}\cdot
        2^{\mathrm{E}_8[\lceil i/2\rceil]}}
        \geq \mathrm{TH}_4
    \right),
    \quad i=1,\ldots,16,
    \label{eq:HiF4_e4}
\end{equation}
where the original threshold is $\mathrm{TH}_4=2$.

\paragraph{Element-wise scaling and S1P2 quantization.}
For each element $v_j$, the effective scale is jointly determined by the global scale and the corresponding two levels of scale extensions:
\begin{equation}
    S_j
    =
    \widehat{\mathrm{SF}}
    \cdot
    2^{
        \mathrm{E}_8[\lceil j/8\rceil]
        +
        \mathrm{E}_4[\lceil j/4\rceil]
    },
    \quad j=1,\ldots,64.
    \label{eq:HiF4_effective_scale}
\end{equation}
Each element is normalized by its effective scale and quantized using
the S1P2 codebook. The S1P2 format allocates one sign bit, two exponent
bits, and one mantissa bit; the sign is carried separately, and the
magnitude codebook contains eight non-negative levels:
\begin{equation}
    \mathcal{Q}_{\mathrm{S1P2}}
    =
    \{0,0.25,0.5,0.75,1.0,1.25,1.5,1.75\}.
    \label{eq:HiF4_codebook}
\end{equation}
The reconstructed element is
\begin{equation}
    \widehat{v}_j
    =
    \operatorname{sign}(v_j)
    \cdot S_j
    \cdot
    Q_{\mathrm{S1P2}}
    \left(
        \frac{|v_j|}{S_j}
    \right),
    \label{eq:HiF4_reconstruction}
\end{equation}
where $Q_{\mathrm{S1P2}}(\cdot)$ returns the nearest element of $\mathcal{Q}_{\mathrm{S1P2}}$ with saturation to the representable range.

The three scale-encoding decisions can therefore be parameterized as
\begin{equation}
    \boldsymbol{\theta}
    =
    (\alpha_{\mathrm{rnd}},\mathrm{TH}_8,\mathrm{TH}_4).
    \label{eq:HiF4_encoding_parameters}
\end{equation}
The original HiF4 configuration uses
\begin{equation}
    \boldsymbol{\theta}_{\mathrm{orig}}
    =
    (0,4,2).
    \label{eq:HiF4_original_parameters}
\end{equation}
Since the Level-1 base scale determines the normalized magnitudes used by both subsequent extension levels, these three encoding parameters jointly influence the effective scales assigned to the 64 elements and, consequently, the final quantization error. This parameterized view provides the basis for investigating and calibrating the hierarchical scale-encoding decisions.

\subsection{Level-1 E6M2 Scale Rounding-Bias Calibration}
\label{subsec:level1_rounding_calibration}

As described in Section~\ref{subsec:HiF4_quantization}, the Level-1 E6M2 scale determines the reference scale for the subsequent hierarchical extensions. Consequently, its rounding decision affects not only the global scale representation, but also the Level-2 and Level-3 scale-extension decisions and the final element-wise quantization error. We therefore investigate whether a small bias in the Level-1 rounding rule can better align the default scale code\cite{nagel2020adaround} with the activation distribution. Specifically, we use the rounding adjustment $\alpha_{\mathrm{rnd}}$ defined in Eq.~\eqref{eq:HiF4_e6m2}, while retaining the original HiF4 hierarchy and thresholds ($\mathrm{TH}_8=4$ and $\mathrm{TH}_4=2$).

To analyse the scale-code alignment, we introduce a small set of ordinal offsets, $\mathcal{F}=\{-1,0,+1,+2\}$, solely for offline evaluation. For each activation block and each candidate offset, we recompute the complete HiF4 quantization--dequantization (QDQ) path, including both hierarchical scale-extension decisions. Let $L_b(\alpha_{\mathrm{rnd}},f)$ denote the resulting block-wise reconstruction MSE. We record the offset that gives the lowest error for each block,
\begin{equation}
    f_b^*(\alpha_{\mathrm{rnd}})
    =\arg\min_{f\in\mathcal{F}} L_b(\alpha_{\mathrm{rnd}},f),
    \qquad
    P_f(\alpha_{\mathrm{rnd}})
    =\frac{1}{B}\sum_{b=1}^{B}
    \mathbb{I}\!\left[f_b^*(\alpha_{\mathrm{rnd}})=f\right].
    \label{eq:l1_offset_distribution}
\end{equation}
Here, $P_0$ is the proportion of blocks for which the unshifted Level-1 code is the best among the four tested candidates. This offset distribution is used to diagnose the alignment between the default scale code and the block-wise preferred codes; the per-block offset selection is not used during inference. The complete offline calibration and offset-analysis procedure, including E6M2 mantissa-range correction and full-QDQ evaluation, is detailed in Appendix~\ref{app:l1_bias_calibration}, Algorithm~\ref{alg:l1_bias_calibration}.

At deployment, we retain the default offset $f=0$ and apply only the fixed rounding adjustment. The corresponding distortion is therefore
\begin{equation}
    L_{\mathrm{fixed}}(\alpha_{\mathrm{rnd}})
    =\frac{1}{B}\sum_{b=1}^{B}
    L_b(\alpha_{\mathrm{rnd}},0),
    \qquad
    \alpha_{\mathrm{rnd}}^*
    =\arg\min_{\alpha_{\mathrm{rnd}}\in\mathcal{A}}
    L_{\mathrm{fixed}}(\alpha_{\mathrm{rnd}}),
    \label{eq:l1_fixed_mse}
\end{equation}
where $\mathcal{A}$ is the set of evaluated rounding adjustments. We also report the restricted four-offset oracle MSE,
\begin{equation}
    L_{\mathrm{oracle}}(\alpha_{\mathrm{rnd}})
    =\frac{1}{B}\sum_{b=1}^{B}
    \min_{f\in\mathcal{F}}L_b(\alpha_{\mathrm{rnd}},f),
    \label{eq:l1_oracle_mse}
\end{equation}
which serves only as an offline reference. The deployed parameter is selected using $L_{\mathrm{fixed}}$, rather than the oracle error or the offset frequency $P_0$.

\begin{figure*}[t]
    \centering

    \begin{subfigure}[t]{0.485\textwidth}
        \centering
        \includegraphics[width=\linewidth]
        {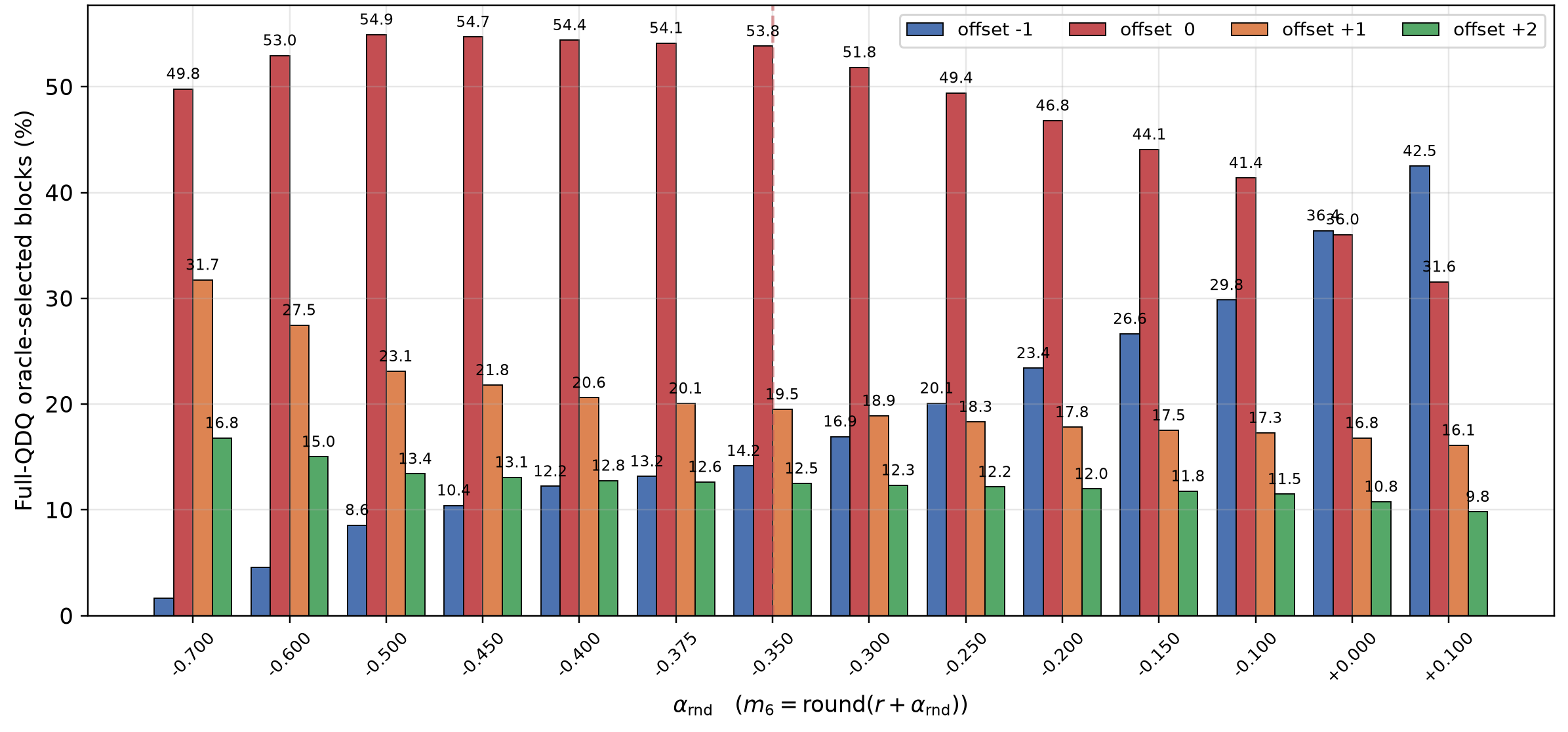}
        \caption{Full-QDQ offset distribution on Qwen3-8B real activations.}
        \label{fig:l1_offset_distribution}
    \end{subfigure}
    \hfill
    \begin{subfigure}[t]{0.485\textwidth}
        \centering
        \includegraphics[width=\linewidth]
        {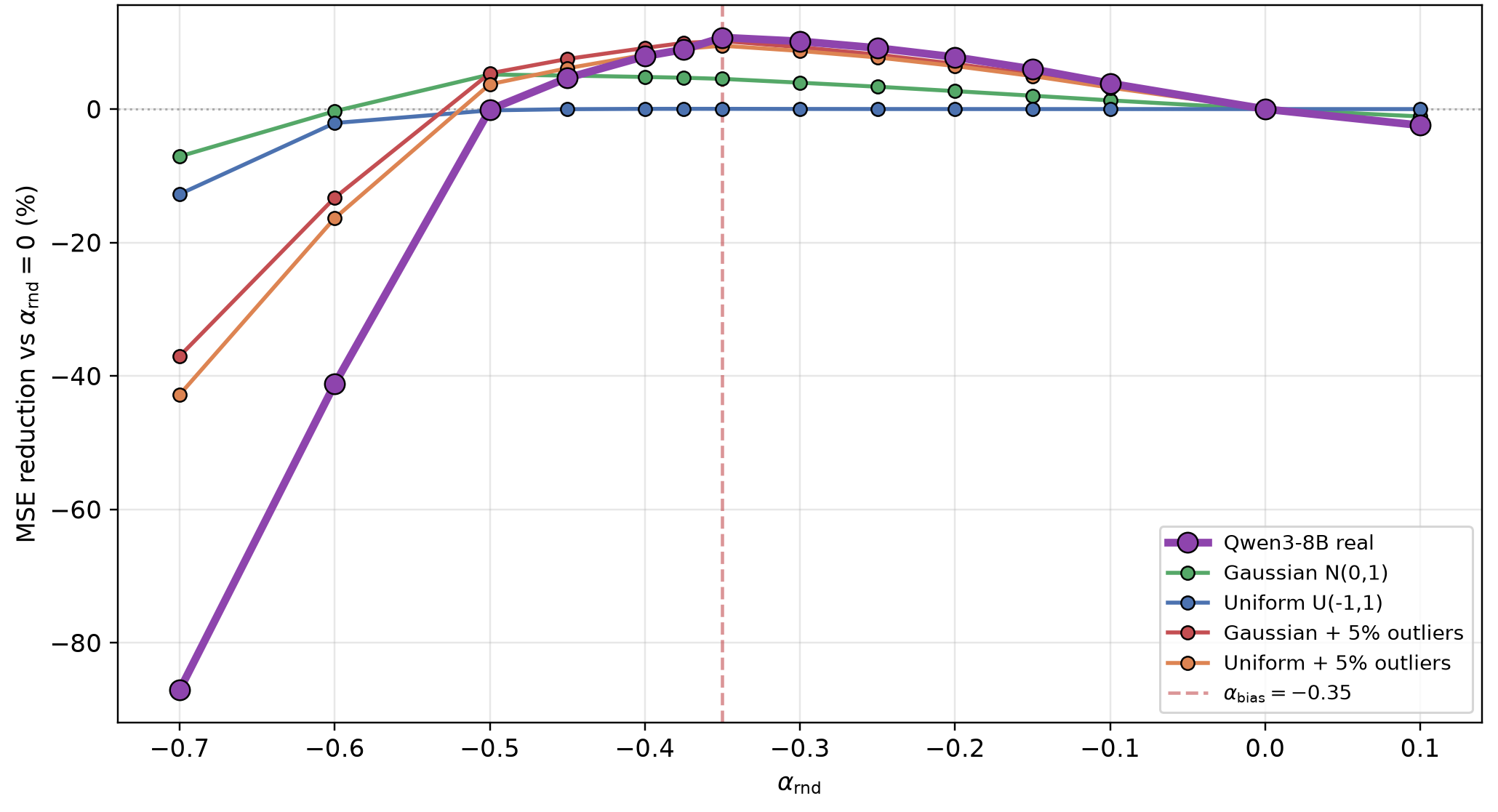}
        \caption{Relative fixed-rule MSE reduction across five activation distributions.}
        \label{fig:l1_relative_mse}
    \end{subfigure}

    \caption{
    Level-1 E6M2 rounding-bias calibration.
    (a) A negative rounding bias shifts the oracle-selected
    offset distribution towards the default offset of zero
    on Qwen3-8B real activations.
    (b) Relative fixed-rule QDQ MSE reduction across Qwen3-8B
    real activations and four synthetic distributions.
    The outlier-bearing distributions exhibit improvements
    comparable to those observed on real activations.
    The dashed vertical line indicates the selected
    $\alpha_{\mathrm{rnd}}=-0.35$ for Qwen3-8B.
    }
    \label{fig:l1_rounding_calibration}
\end{figure*}

Figure~\ref{fig:l1_rounding_calibration} first illustrates the
offset distribution on Qwen3-8B real activation blocks.
As shown in Fig.~\ref{fig:l1_rounding_calibration}(a),
the original setting ($\alpha_{\mathrm{rnd}}=0$) exhibits
a considerable mismatch between the default scale code and
the block-wise preferred codes: only $36.03\%$ of blocks
select offset zero, whereas $36.40\%$ select offset $-1$.
Introducing a negative rounding bias shifts the offset
distribution towards zero. In particular, at
$\alpha_{\mathrm{rnd}}=-0.35$, the proportion selecting
offset zero increases to $53.83\%$, while the proportion
selecting offset $-1$ decreases to $14.16\%$.
This indicates improved alignment between the default
Level-1 scale code and the preferred codes under the
complete hierarchical QDQ process.

To further investigate the sensitivity of the proposed calibration
to activation distributions, Fig.~\ref{fig:l1_rounding_calibration}(b)
compares the relative fixed-rule MSE changes across Qwen3-8B real
activations and four synthetic distributions. Notably, both
synthetic distributions with outliers exhibit substantial
quantization-error reductions at $\alpha_{\mathrm{rnd}}=-0.35$:
$10.41\%$ for Gaussian activations with $5\%$ outliers and
$9.62\%$ for uniform activations with $5\%$ outliers.
These improvements are comparable to the $10.93\%$ reduction
observed on Qwen3-8B real activations. In contrast, the
improvement is much smaller for Gaussian activations without
outliers ($4.59\%$), while the uniform distribution exhibits
the largest reduction ($22.75\%$) among the four synthetic
distributions. This observation suggests that the
rounding-bias calibration is particularly beneficial for
the evaluated outlier-bearing distributions, which is consistent
with the long-tailed and outlier-bearing characteristics
motivating quantization calibration for real LLM activations.

Nevertheless, the optimal rounding adjustment remains
distribution-dependent. While $\alpha_{\mathrm{rnd}}=-0.35$
achieves the lowest fixed-rule MSE among the evaluated candidates
for Qwen3-8B and both synthetic outlier distributions, the
Gaussian distribution favors $\alpha_{\mathrm{rnd}}=-0.50$, and
the uniform distribution's reduction is essentially flat for
$\alpha_{\mathrm{rnd}}\in[-0.50,-0.10]$ (remaining at
approximately $22.75\%$ throughout this range). We therefore
select $\alpha_{\mathrm{rnd}}=-0.35$ according to the minimum
fixed-rule MSE on Qwen3-8B real activations, rather than assuming
a universally optimal rounding bias.

\subsection{Hierarchical Scale-Extension Threshold Calibration}
\label{subsec:hierarchical_threshold_calibration}

Following the Level-1 E6M2 scale calibration in Section~\ref{subsec:level1_rounding_calibration},
we calibrate the Level-2 and Level-3 scale-extension thresholds. The threshold calibration is applied downstream of the Level-1 rounding bias: the hierarchical state labels used in the KDE analysis are derived from the biased scale $\widehat{\mathrm{SF}}$ at $\alpha_{\mathrm{rnd}}=-0.35$. The same threshold pair is also selected at $\alpha_{\mathrm{rnd}}=0$ within the evaluated search grid, suggesting that the selected thresholds are robust to the tested Level-1 rounding settings.

\begin{figure}[t]
    \centering
    \includegraphics[width=0.98\linewidth]{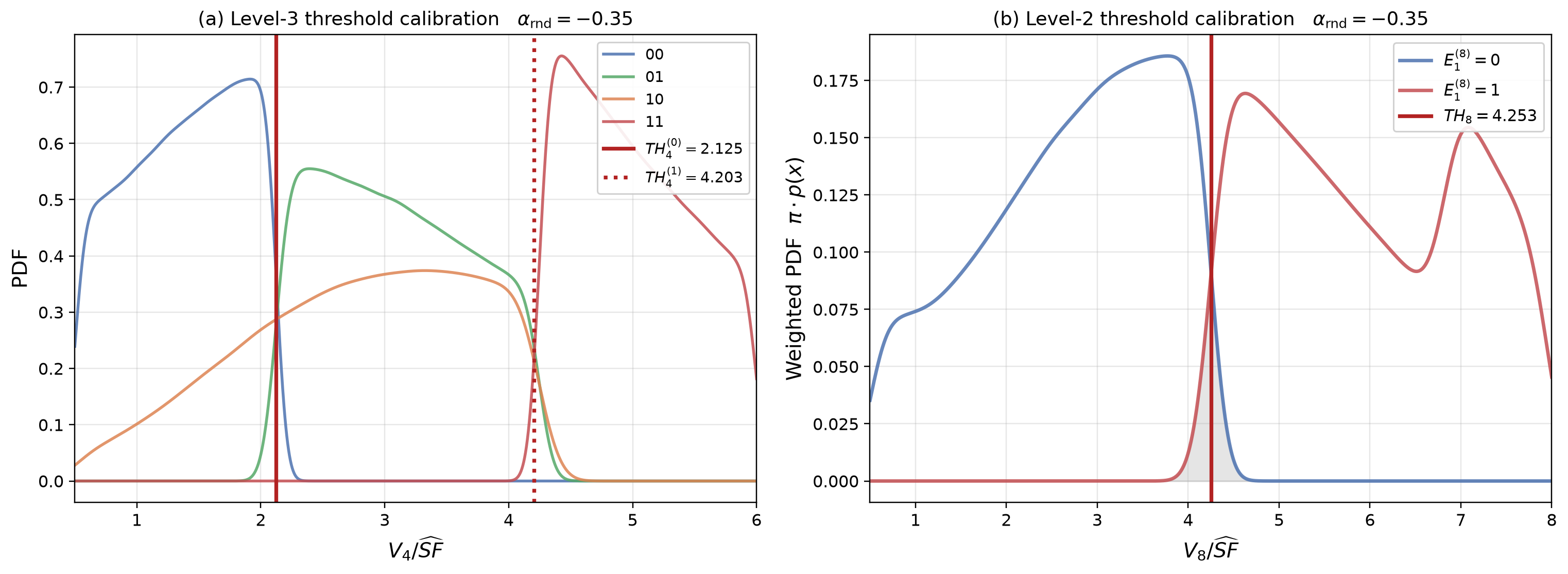}
    \caption{KDE-based calibration of the HiF4 hierarchical thresholds on
    Qwen3-8B FFN-input activations with $\alpha_{\mathrm{rnd}}=-0.35$.
    (a) Level-3 state distributions and conditional crossovers: $2.1246$
    for $E_8=0$, and $4.2035$ on the raw coordinate for $E_8=1$
    ($2.10175$ after normalization). (b) Level-2 parent-state distributions,
    with a crossover at $4.2529$. The selected thresholds are
    $TH_4=2.125$ and $TH_8=4.25$.}
    \label{fig:sec23_optimal_thresholds}
\end{figure}

\paragraph{Optimal hierarchical states.}
Each eight-element parent shares one Level-2 bit $E_8$ between its two
four-element children, each with its own Level-3 bit $E_4$. As defined
previously, the effective extension is $DE=E_8+E_4$. Let $\mathcal{L}_i(DE)$ denote the reconstruction MSE of the
$i$-th four-element group when using the effective scale
$\widehat{\mathrm{SF}}2^{DE}$:
\begin{equation}
    \mathcal{L}_i(DE)
    =
    \frac{1}{4}\sum_{j=4i-3}^{4i}
    \left(v_j-\widehat{v}_j(DE)\right)^2,
\end{equation}
where $\widehat{v}_j(DE)$ is obtained using the original
S1P2 reconstruction in Eq.~\eqref{eq:HiF4_reconstruction}.
We jointly select the
shared parent bit and the two child bits by minimizing the sum of the two
child-group reconstruction losses over the eight legal assignments:
\begin{equation}
\left(E_8^*,E_{4,1}^*,E_{4,2}^*\right)
=\arg\min_{E_8,E_{4,1},E_{4,2}\in\{0,1\}}
\sum_{j=1}^{2}\mathcal{L}_{i_j}\!\left(E_8+E_{4,j}\right).
\label{eq:sec23_optimal_hierarchical_state}
\end{equation}
For the $k$-th eight-element parent group, its two child groups
are indexed by $i_1=2k-1$ and $i_2=2k$.
The resulting child labels are $00$, $01$, $10$, and $11$, and are used as
class labels in the subsequent KDE analysis.

\paragraph{KDE-based threshold estimation.}
We use 200,000 randomly sampled 64-element blocks from Qwen3-8B
activations. For both extension levels, the KDE inputs are the
normalized magnitude statistics
\begin{equation}
r_8 = \frac{V_8}{\widehat{SF}}, \qquad
r_4 = \frac{V_4}{\widehat{SF}\cdot 2^{E_8}}.
\label{eq:sec23_kde_features}
\end{equation}
Based on the optimal hierarchical state labels, KDE is used to
estimate the intersections between the corresponding distributions
of $r_8$ and $r_4$. For Level-2, we compare the two parent
classes $E_8=0$ and $E_8=1$ using $r_8$. For Level-3, we compare
states $00$ and $01$ under $E_8=0$, and states $10$ and $11$ under
$E_8=1$, using the corresponding $r_4$ values.

As shown in Fig.~\ref{fig:sec23_optimal_thresholds}, the Level-2 crossover
is $4.2529$, approximately $4.25$. The two Level-3 crossovers are $2.1246$
for $E_8=0$ and $4.2035$ on the raw coordinate $u_4=V_4/\widehat{SF}$ for
$E_8=1$. Since $r_4=u_4/2$ when $E_8=1$, the latter corresponds to
$4.2035/2=2.10175$ in the normalized coordinate. Both Level-3 estimates are
close to $2.125$. We therefore select
\begin{equation}
    TH_8=4.25,\qquad TH_4=2.125,
    \label{eq:sec23_selected_thresholds}
\end{equation}
which satisfy the hierarchical relation $TH_8 = 2TH_4$ and are dyadic
rationals ($4.25 = 17/4$, $2.125 = 17/8$), so the comparisons
$V_8/\widehat{SF} \geq 4.25$ and $V_4/(\widehat{\mathrm{SF}}\cdot 2^{E_8}) \geq 2.125$
can be implemented as integer comparisons
$4V_8 \geq 17\widehat{\mathrm{SF}}$ and $8V_4 \geq 17\widehat{\mathrm{SF}}\cdot 2^{E_8}$.

\begin{figure}[t]
    \centering
    \includegraphics[width=0.77\linewidth]{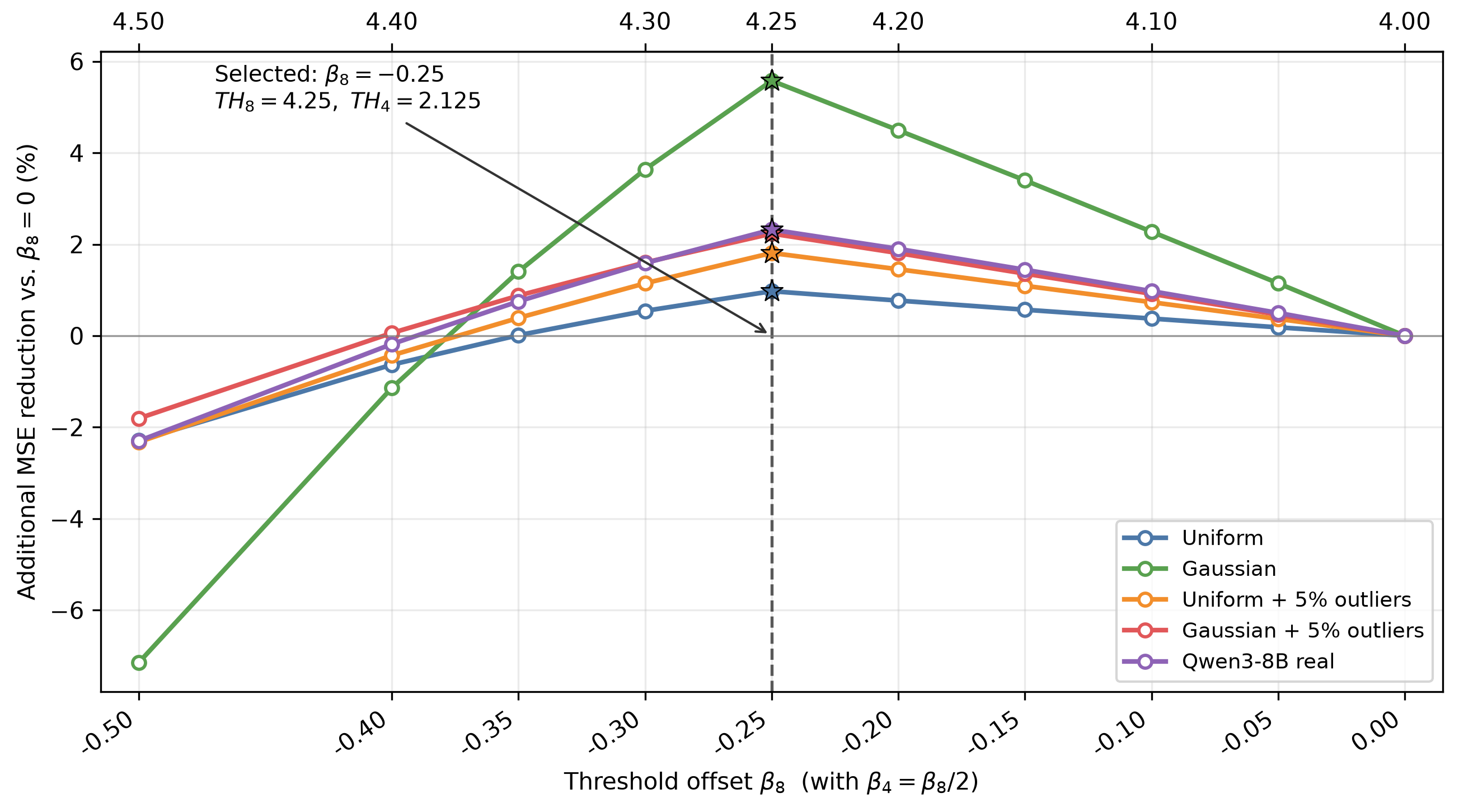}
    \caption{HiF4 QDQ threshold sweep across five activation
    distributions with $\alpha_{\mathrm{rnd}}=-0.35$. The vertical marker
    indicates $\beta_8=-0.25$, corresponding to
    $TH_8=4.25$ and $TH_4=2.125$. All five distributions achieve their
    largest MSE reduction at this setting.}
    \label{fig:sec23_beta_sweep}
\end{figure}

\paragraph{End-to-end QDQ validation.}
We further validate the selected thresholds using the complete HiF4
quantize--dequantize (QDQ) path across five activation distributions, while
preserving $TH_8=2TH_4$. The thresholds are parameterized as
\begin{equation}
    TH_8=4-\beta_8,\qquad
    TH_4=2-\frac{\beta_8}{2},
    \label{eq:sec23_beta_parameterization}
\end{equation}
and evaluated over
$\beta_8\in\{0,-0.05,-0.10,-0.15,-0.20,-0.25,-0.30,-0.35,-0.40,-0.50\}$.
For each distribution, the additional relative MSE reduction is measured
against its own result at the original thresholds ($\beta_8=0$):
\begin{equation}
    \Delta_{\mathrm{MSE}}(\beta_8)
    =\frac{\mathcal{L}_{\mathrm{QDQ}}(0)
    -\mathcal{L}_{\mathrm{QDQ}}(\beta_8)}
    {\mathcal{L}_{\mathrm{QDQ}}(0)}\times100\%.
    \label{eq:sec23_beta_gain}
\end{equation}

Figure~\ref{fig:sec23_beta_sweep} shows that all five distributions achieve
their largest MSE reduction at $\beta_8=-0.25$, corresponding to
$(TH_8,TH_4)=(4.25,2.125)$. This agrees with the KDE-based threshold
estimates. For Qwen3-8B, the QDQ MSE decreases from
$2.181089\times10^{-3}$ to $2.130460\times10^{-3}$, an additional reduction
of approximately $2.32\%$.



\section{Proposed HiF4 Quantization Algorithm}
\label{sec:proposed_HiF4_algorithm}

Based on the calibration results in Sections~\ref{subsec:level1_rounding_calibration} and
\ref{subsec:hierarchical_threshold_calibration}, we instantiate the HiF4 quantizer with three fixed parameters:
$\alpha_{\mathrm{rnd}}=-0.35$, $TH_8=4.25$, and $TH_4=2.125$.
The procedure follows the original data organization and reconstruction
formulation in Section~\ref{subsec:HiF4_quantization}. In particular, the
64-element block is partitioned into sixteen 4-element groups and eight
8-element parent groups, whose peak values $V_4[i]$, $V_8[k]$, and
$V_{\max}$ are defined in Eqs.~\eqref{eq:HiF4_v4}--\eqref{eq:HiF4_vmax}.
The S1P2 element codebook and reconstruction operation in
Eq.~\eqref{eq:HiF4_reconstruction} are kept unchanged.

The proposed method introduces three modifications to the scale-control
parameters of the original HiF4 quantization path. At Level-1, we introduce
a fixed rounding bias of $\alpha_{\mathrm{rnd}}=-0.35$ instead of the
original unbiased setting ($\alpha_{\mathrm{rnd}}=0$). At Level-2 and
Level-3, we adjust the scale-extension thresholds to $TH_8=4.25$ and
$TH_4=2.125$, respectively. The remaining quantization operations follow
the original HiF4 path.

Algorithm~\ref{alg:final_HiF4_quantization} gives the corresponding
block-wise procedure using the definitions and reconstruction equations
from Section~\ref{subsec:HiF4_quantization}. Inline comments distinguish the
three proposed changes from the operations retained from the original HiF4
path.

\begin{algorithm}[t]
    \caption{Final Deterministic HiF4 Quantization for a 64-Element Block}
    \label{alg:final_HiF4_quantization}
    \begin{algorithmic}[1]
        \Require Input block $X=\{x_j\}_{j=1}^{64}$
        \Ensure Quantized S1P2 codes $\{(\sigma_j,q_j)\}_{j=1}^{64}$ and scale metadata
        \State Compute $V_4[i]$, $V_8[k]$, and $V_{\max}$ as in Eqs.~\eqref{eq:HiF4_v4}--\eqref{eq:HiF4_vmax} \Comment{Original grouping and peak extraction}
        \State $SF\gets\operatorname{clamp}(V_{\max}/7,2^{-48},49152)$ \Comment{Original HiF4 block-scale rule}
        \State $E_{SF}\gets\lfloor\log_2(SF)\rfloor$; $r\gets SF\,2^{2-E_{SF}}$
        \State $m_{\mathrm{raw}}\gets\operatorname{round}(r-0.35)$
            \Comment{\textbf{Modified:} Level-1 biased rounding, $\alpha_{\mathrm{rnd}}=-0.35$}
        \State $E6M2\gets\operatorname{Encode}_{E6M2}(E_{SF},m_{\mathrm{raw}})$
            \Comment{Apply the original E6M2 encoding and range handling}
        \State $(m_{\mathrm{can}}, E_{\mathrm{can}}) \gets \operatorname{Decode}_{E6M2}(E6M2)$
        \State $\widehat{SF} \gets m_{\mathrm{can}}\,2^{E_{\mathrm{can}}-2}$
        \For{$k=1$ to $8$} \Comment{\textbf{Modified:} $TH_8=4.25$ (original: $4$)}
            \State $e_{8,k}\gets\mathbb{I}\!\left(V_8[k]/\widehat{SF}\geq4.25\right)$
        \EndFor
        \For{$i=1$ to $16$} \Comment{\textbf{Modified:} $TH_4=2.125$ (original: $2$)}
            \State $k\gets\lceil i/2\rceil$ \Comment{Two children share one parent bit}
            \State $e_{4,i}\gets\mathbb{I}\!\left(V_4[i]/(\widehat{SF}\,2^{e_{8,k}})\geq2.125\right)$
            \State $DE_i\gets e_{8,k}+e_{4,i}$; $S_i\gets\widehat{SF}\,2^{DE_i}$ \Comment{Combine extensions to obtain the effective scale}
        \EndFor
        \For{$i=1$ to $16$} \Comment{Original S1P2 element quantization}
            \For{each element $x_j\in\mathcal{G}_i$}
                \State $\sigma_j\gets\operatorname{sign}(x_j)$; $q_j\gets Q_{\mathrm{S1P2}}(|x_j|/S_i)$
            \EndFor
        \EndFor
        \State \Return $\{(\sigma_j,q_j)\}_{j=1}^{64}$ and metadata
        $(E6M2,\{e_{8,k}\}_{k=1}^{8},\{e_{4,i}\}_{i=1}^{16})$
    \end{algorithmic}
\end{algorithm}

After mantissa-range correction and the original E6M2 range handling, the resulting
$\widehat{SF}$ is used by both hierarchy levels. The effective extension and
scale $DE_i$ and $S_i$ are exactly those in
Eq.~\eqref{eq:HiF4_effective_scale}; the output codes are reconstructed by
Eq.~\eqref{eq:HiF4_reconstruction}. Therefore, the proposed method changes
only the three calibrated scale-control parameters and retains the original
HiF4 grouping, metadata hierarchy, and element-wise codebook.

\begin{table*}[t]
    \centering
    \small
    \setlength{\tabcolsep}{5pt}
    \caption{QDQ MSE comparison of different HiF4 configurations across
    five activation distributions. All MSE values are reported in units
    of $10^{-3}$. Each column header shows the parameter tuple
    $(\alpha_{\mathrm{rnd}},TH_8,TH_4)$. Parenthetical values indicate
    relative MSE reductions compared with Original HiF4, calculated
    using unrounded MSE values.}
    \label{tab:HiF4_mse_five_distributions}
    \begin{tabular}{lccc}
        \toprule
        Distribution
        & \makecell[c]{Original HiF4\\$(0,4,2)$}
        & \makecell[c]{Bias-only\\$(-0.35,4,2)$}
        & \makecell[c]{Proposed\\$(-0.35,4.25,2.125)$} \\
        \midrule
        Uniform
        & 1.60
        & 1.24 (22.75\%)
        & 1.23 (23.50\%) \\
        Gaussian
        & 2.19
        & 2.09 (4.59\%)
        & 1.97 (9.91\%) \\
        Uniform + outliers
        & 3.61
        & 3.27 (9.62\%)
        & 3.21 (11.25\%) \\
        Gaussian + outliers
        & 3.51
        & 3.15 (10.41\%)
        & 3.08 (12.41\%) \\
        Qwen3-8B activations
        & 2.45
        & 2.18 (10.93\%)
        & 2.13 (13.00\%) \\
        \bottomrule
    \end{tabular}
\end{table*}

Table~\ref{tab:HiF4_mse_five_distributions} compares the QDQ MSE
of the original HiF4 configuration, the bias-only configuration,
and the proposed configuration across five activation distributions.
The comparison separates the contribution of Level-1 biased rounding
from that of the hierarchical threshold adjustments.
Specifically, the difference between Original HiF4 and Bias-only
reflects the effect of introducing
$\alpha_{\mathrm{rnd}}=-0.35$, while the difference between
Bias-only and Proposed measures the additional benefit of adjusting
$TH_8$ and $TH_4$.
The results demonstrate the effectiveness of combining the three
calibrated parameters in the final HiF4 quantization method.

\section{Accuracy Evaluation and Results}
\label{sec:experiments}

We evaluate three numerical configurations---FP16 reference, original HiF4, and the proposed HiF4 Scale-bias method---on Qwen3-8B and Qwen3-30B-A3B \cite{qwen3} across six benchmarks: MMLU \cite{mmlu}, GSM8K \cite{gsm8k}, HumanEval \cite{humaneval}, MBPP \cite{mbpp}, ARC-Challenge (ARC-C) and ARC-Easy (ARC-E) \cite{arc}. To reduce measurement noise, each reported score is averaged over multiple independent evaluation runs, and the Mean column reports the arithmetic average over the six benchmark means. In all quantized configurations, both weight and activation tensors are quantized with the HiF4 format using the same scale-encoding parameters; the FP16 reference is used only as the accuracy baseline.

At the quantization level (Table~\ref{tab:HiF4_mse_five_distributions}),
the proposed configuration reduces QDQ MSE by 23.50\% for Uniform,
9.91\% for Gaussian, 11.25\% for Uniform with outliers, 12.41\% for
Gaussian with outliers, and 13.00\% on Qwen3-8B real activations,
relative to original HiF4. These reductions decompose into a Level-1
rounding-bias contribution and an additional hierarchical-threshold
contribution from $(TH_8,TH_4)=(4.25,2.125)$, indicating that the
calibrated parameters lower tensor-level reconstruction error across
all five evaluated distributions.

Table~\ref{tab:accuracy_qwen3} reports the end-to-end benchmark results.
For Qwen3-8B, the six-benchmark mean improves from 72.29\% under
original HiF4 to 73.20\% under the proposed method, reducing the FP16
gap from 2.31\,pp to 1.40\,pp. For Qwen3-30B-A3B, the mean improves
from 79.01\% to 80.26\%, reducing the FP16 gap from 2.39\,pp to
1.14\,pp. The proposed method improves the six-benchmark mean for both
evaluated model sizes and reduces the accuracy loss relative to their
FP16 references. The QDQ and benchmark results are complementary. The calibrated
parameters reduce reconstruction error across synthetic and real
activations, and this reduction is accompanied by improved six-task
mean accuracy on both evaluated models.

\begin{table*}[t]
    \centering
    \small
    \setlength{\tabcolsep}{5pt}
    \caption{End-to-end accuracy (\%) of Qwen3 models under FP16, original
    HiF4, and the proposed HiF4 Scale-bias method. The Mean column reports
    the arithmetic mean over six benchmarks. Values in parentheses indicate
    the difference from the corresponding FP16 mean, in percentage points.}
    \label{tab:accuracy_qwen3}
    \begin{tabular}{llrrrrrrr}
        \toprule
        \textbf{Model} & \textbf{Method} & \textbf{MMLU} &
        \textbf{GSM8K} & \textbf{HumanEval} & \textbf{MBPP} &
        \textbf{ARC-C} & \textbf{ARC-E} & \textbf{Mean} \\
        \midrule
        \multirow{3}{*}{Qwen3-8B}
        & FP16 & 74.97 & 88.48 & 63.62 & 65.00 & 67.58 & 87.96 & 74.60 \\
        & HiF4 & 73.16 & 85.14 & 59.15 & 63.20 & 65.53 & 87.54 & 72.29 (-2.31) \\
        & HiF4 Scale-bias & 72.90 & 86.05 & 65.85 & 62.60 & 64.76 & 87.04 & \textbf{73.20 (-1.40)} \\
        \midrule
        \multirow{3}{*}{Qwen3-30B-A3B}
        & FP16 & 82.03 & 89.82 & 76.42 & 76.73 & 72.81 & 90.56 & 81.40 \\
        & HiF4 & 80.79 & 88.80 & 69.82 & 73.80 & 71.08 & 89.76 & 79.01 (-2.39) \\
        & HiF4 Scale-bias & 81.14 & 88.60 & 74.49 & 75.13 & 72.04 & 90.15 & \textbf{80.26 (-1.14)} \\
        \bottomrule
    \end{tabular}
\end{table*}

\subsection{Hardware Friendliness}
\label{subsec:hardware_friendliness}

The three calibrated parameters refine the HiF4 scale-encoding rules without requiring new hardware operations. At the instruction level,
the Level-1 rounding bias is realized by a single constant added to
the normalized E6M2 mantissa before the existing rounding step, and
the Level-2 and Level-3 thresholds replace the immediate constants in
the existing magnitude comparisons. Both operations are covered by the
floating-point add, multiply, compare, and rounding instructions that
are already present in general-purpose inference pipelines; no new
instruction category, operand format, or data path is required.

The three calibrated values ($\alpha_{\mathrm{rnd}} = -0.35$,
$TH_8 = 4.25$, and $TH_4 = 2.125$) are compile-time constants and do
not require additional runtime storage beyond the existing HiF4 scale
metadata. Only customized instruction extensions or fused operators
that hard-wire the original rounding constant or threshold values may
require minor adjustment to expose the new parameters, and even in
that case the change is limited to substituting the affected constants
rather than redesigning the arithmetic pipeline; this is consistent
with recent HiF4 deployment efforts on Ascend NPUs
\cite{dong2026hifa4}.

This single-path property distinguishes the proposed method from
per-block scale-search approaches such as ScaleSearch
\cite{gupta2026scalesearch}, which evaluate multiple candidate scales
for each block and therefore require one quantization path per
candidate. In a hardware implementation, such per-block search either
replicates the quantization datapath or serializes the candidate
evaluations, both of which increase area or latency proportionally to
the number of candidates. The proposed method avoids this cost
entirely: all three scale-encoding parameters are fixed after offline
calibration, and each block is quantized through a single
deterministic path.

\subsection{Discussion}
\label{subsec:accuracy_discussion}

The quantization-level and end-to-end results provide complementary
evidence for the proposed scale-calibration method, and the hardware
implications discussed above confirm that these accuracy gains do not
require new arithmetic support. The calibrated parameters reduce
reconstruction error on synthetic distributions and real Qwen3-8B
activations, and this reduction is accompanied by improved six-task
mean accuracy on both Qwen3-8B and Qwen3-30B-A3B. A lower tensor-level
MSE does not necessarily translate into an equal improvement on every
benchmark, because the six benchmarks have different sensitivities to
quantization noise and different inherent evaluation variance; the
consistent mean-accuracy gains across two model sizes nevertheless
indicate that the calibration is useful beyond the local QDQ metric.

The proposed method occupies a different point in the accuracy--cost
trade-off than per-block scale-search approaches such as ScaleSearch
\cite{gupta2026scalesearch}. ScaleSearch improves quantization
accuracy by searching over neighboring representable scales for each
block, at the cost of evaluating multiple quantization candidates at
inference time. The proposed method instead recovers most of the
accuracy gain achievable by per-block scale selection while fixing the
encoding parameters to a single calibration-derived value, so that the
deployed quantizer retains a deterministic single-pass structure. This
trade-off is particularly relevant for hardware accelerators, where
the cost of per-block multi-candidate evaluation scales with the
number of candidates rather than being amortized across blocks.

The method retains the original HiF4 hierarchical grouping and S1P2
element codebook. Its changes are limited to the Level-1 rounding bias
and the Level-2/Level-3 scale-extension thresholds; these are selected
through offline calibration and applied as fixed rules during
quantization, without per-block enumeration of multiple scale
candidates. This keeps the quantization procedure deterministic and
preserves the existing HiF4 representation structure. The
transferability of the calibrated parameter set to other model
architectures and activation distributions remains to be studied.

\section{Conclusion}
\label{sec:conclusion}
This paper presents a deterministic scale-calibration method that improves HiF4 quantization accuracy by refining the scale-encoding decisions at all three levels. The final configuration uses $\alpha_{\mathrm{rnd}}=-0.35$, $TH_8=4.25$, and
$TH_4=2.125$. These parameters are determined through offline calibration
and are applied directly during quantization, without per-block enumeration
of multiple scale candidates. The original HiF4 hierarchical organization
and S1P2 element codebook are retained.

The quantization-level evaluation demonstrates that the combined calibration
reduces QDQ MSE across five tested activation distributions, including real
Qwen3-8B activations. The end-to-end evaluation further shows that the
six-benchmark mean accuracy improves from 72.29\% to 73.20\% for Qwen3-8B
and from 79.01\% to 80.26\% for Qwen3-30B-A3B. Correspondingly, the accuracy
gaps to the FP16 references decrease from 2.31 pp to 1.40 pp and from
2.39 pp to 1.14 pp, respectively. These results indicate that calibrating
the hierarchical scale representation can improve the average end-to-end
accuracy of HiF4 across the evaluated models.

\section*{Acknowledgements}
AI-based tools were used to assist with language editing and manuscript preparation. The authors take full responsibility for the technical content, methodology, experiments, and conclusions presented in this paper.

\bibliographystyle{unsrt}  
\bibliography{references}  

\clearpage
\begin{appendices}
\section{Implementation of Level-1 Rounding-Bias Calibration}
\label{app:l1_bias_calibration}

Algorithm~\ref{alg:l1_bias_calibration} summarizes the offline
calibration procedure. For each candidate rounding bias, the
normalized E6M2 mantissa is rounded and adjusted to the valid
mantissa range following the Level-1 encoding procedure in
Section~\ref{subsec:HiF4_quantization}. Neighboring ordinal scale
codes are then evaluated using the complete HiF4 QDQ path,
including the two hierarchical scale-extension decisions and
the original S1P2 reconstruction. The offset distribution is
used only to analyze scale-code alignment, whereas the deployed
bias is selected using the fixed-rule MSE with zero ordinal
offset.

\begin{algorithm}[H]
\caption{Offline Level-1 Rounding-Bias Calibration and Offset Analysis}
\label{alg:l1_bias_calibration}
\begin{algorithmic}[1]

\Require Activation blocks $\mathcal{D}=\{\mathbf{x}_b\}_{b=1}^{B}$;
         candidate biases $\mathcal{A}$;
         ordinal offsets $\mathcal{F}=\{-1,0,1,2\}$;
         thresholds $(TH_8,TH_4)=(4,2)$
\Ensure Selected bias $\alpha^*$; fixed-rule MSE
        $L_{\mathrm{fixed}}(\alpha^*)$; offset distributions
        $\{P_f(\alpha)\}$; oracle MSE values
        $\{L_{\mathrm{oracle}}(\alpha)\}$

\For{each block $\mathbf{x}_b\in\mathcal{D}$}
    \State Compute $V_{4,b}[i]$, $V_{8,b}[k]$, and $V_{\max,b}$
    \State $SF_b\gets\operatorname{clamp}(V_{\max,b}/7,2^{-48},49152)$
    \State $E_b\gets\lfloor\log_2 SF_b\rfloor$;
           $r_b\gets SF_b\,2^{2-E_b}$
\EndFor

\For{each $\alpha\in\mathcal{A}$}
    \State Initialize $A_{\mathrm{fixed}}\gets0$,
           $A_{\mathrm{oracle}}\gets0$, and $C_f\gets0$
           for all $f\in\mathcal{F}$

    \For{each block $\mathbf{x}_b\in\mathcal{D}$}
        \Comment{Biased rounding and E6M2 mantissa adjustment}
        \State $m_{6,b}\gets\operatorname{round}(r_b+\alpha)$
        \If{$m_{6,b}<4$}
            \State $(E_{\mathrm{can},b},m_{\mathrm{can},b})\gets
                   (E_b-1,2m_{6,b})$
        \ElsIf{$m_{6,b}\geq8$}
            \State $(E_{\mathrm{can},b},m_{\mathrm{can},b})\gets
                   (E_b+1,m_{6,b}-4)$
        \Else
            \State $(E_{\mathrm{can},b},m_{\mathrm{can},b})\gets
                   (E_b,m_{6,b})$
        \EndIf
        \State $c_b^\alpha\gets
               4(E_{\mathrm{can},b}+48)+(m_{\mathrm{can},b}-4)$

        \For{each $f\in\mathcal{F}$}
            \Comment{Decode a neighboring E6M2 ordinal code}
            \State $c_{b,f}^\alpha\gets
                   \operatorname{clip}(c_b^\alpha+f,0,255)$
            \State $E_{b,f}^\alpha\gets
                   \lfloor c_{b,f}^\alpha/4\rfloor-48$
            \State $M_{b,f}^\alpha\gets
                   4+(c_{b,f}^\alpha\bmod4)$
            \State $\widetilde{SF}_{b,f}^\alpha\gets
                   M_{b,f}^\alpha 2^{E_{b,f}^\alpha-2}$

            \Comment{Apply the original HiF4 hierarchy and S1P2 QDQ}
            \State $\widehat{\mathbf{x}}_{b}^{\alpha,f}\gets
                   \operatorname{QDQ}_{\mathrm{HiF4}}(
                   \mathbf{x}_b;\widetilde{SF}_{b,f}^\alpha,TH_8,TH_4)$
            \State $\mathcal{L}_b(\alpha,f)\gets
                   \frac{1}{64}\left\|
                   \mathbf{x}_b-\widehat{\mathbf{x}}_{b}^{\alpha,f}
                   \right\|_2^2$
        \EndFor

        \Comment{Accumulate offset statistics and reconstruction errors}
        \State $f_b^*(\alpha)\gets
               \arg\min_{f\in\mathcal{F}}\mathcal{L}_b(\alpha,f)$
        \State $C_{f_b^*(\alpha)}\gets C_{f_b^*(\alpha)}+1$
        \State $A_{\mathrm{fixed}}\gets
               A_{\mathrm{fixed}}+\mathcal{L}_b(\alpha,0)$
        \State $A_{\mathrm{oracle}}\gets A_{\mathrm{oracle}}+
               \min_{f\in\mathcal{F}}\mathcal{L}_b(\alpha,f)$
    \EndFor

    \State $L_{\mathrm{fixed}}(\alpha)\gets A_{\mathrm{fixed}}/B$;
           $L_{\mathrm{oracle}}(\alpha)\gets A_{\mathrm{oracle}}/B$
    \For{each $f\in\mathcal{F}$}
        \State $P_f(\alpha)\gets C_f/B$
    \EndFor
\EndFor

\State $\alpha^*\gets\arg\min_{\alpha\in\mathcal{A}}
       L_{\mathrm{fixed}}(\alpha)$
\State \Return $\alpha^*$, $L_{\mathrm{fixed}}(\alpha^*)$,
       $\{P_f(\alpha)\}$, and $\{L_{\mathrm{oracle}}(\alpha)\}$

\end{algorithmic}
\end{algorithm}

Here, $\operatorname{QDQ}_{\mathrm{HiF4}}$ denotes the original
HiF4 quantize--dequantize procedure defined by
Eqs.~\eqref{eq:HiF4_e8}--\eqref{eq:HiF4_reconstruction}.
The fixed-rule MSE uses only the zero offset, whereas the
oracle MSE and $P_f$ are used only for offline analysis.
\end{appendices}

\end{document}